\documentclass[letterpaper, 10 pt, journal, twoside]{ieeetran}

\IEEEoverridecommandlockouts                              

\usepackage{times}

\usepackage{multicol}

\usepackage{acro}
\DeclareAcronym{NCCR}{
  short = NCCR,
  long  = National Center of Competence in Research,
  short-indefinite = an,
  long-indefinite = a
}

\DeclareAcronym{SLF}{
  short = SLF,
  long  = WSL-Institute for Snow and Avalanche Research,
  short-indefinite = an
}

\DeclareAcronym{sUAS}{
  short = sUAS,
  long  = small uncrewed aerial system,
  short-indefinite = an,
  long-indefinite = a
}

\DeclareAcronym{FMU}{
  short = FMU,
  long  = flight management unit,
  short-indefinite = a,
  long-indefinite = a
}

\DeclareAcronym{GNSS}{
  short = GNSS,
  long  = global navigation satellite system,
  short-indefinite = a,
  long-indefinite = a
}

\DeclareAcronym{VTOL}{
  short = VTOL,
  long  = vertical takeoff and landing,
  short-indefinite = v,
  long-indefinite = v
}

\DeclareAcronym{ICS}{
  short = ICS,
  long  = inevitable collision state,
  short-indefinite = an,
  long-indefinite = an
}

\DeclareAcronym{MILP}{
  short = MILP,
  long  = mixed-integer linear programming,
  short-indefinite = a,
  long-indefinite = a
}

\DeclareAcronym{GIS}{
  short = GIS,
  long  = geographic information system,
  short-indefinite = a,
  long-indefinite = a
}

\DeclareAcronym{SITL}{
  short = SITL,
  long  = software-in-the-loop,
  short-indefinite = an,
  long-indefinite = a
}

\DeclareAcronym{RoC}{
  short = RoC,
  long  = rate of climb,
  short-indefinite = an,
  long-indefinite = a
}

\DeclareAcronym{DEM}{
  short = DEM,
  long  = digital elevation map
}

\DeclareAcronym{OMPL}{
  short = OMPL,
  long  = Open Motion Planning Library
}

\DeclareAcronym{ROS}{
  short = ROS,
  long  = Robot Operating System,
  short-indefinite = a
}
\newcommand{\reffig}[1]{Fig.~\ref{#1}}

\newcommand{\refsec}[1]{Section~\ref{#1}}

\newcommand{\refequ}[1]{Eq.~\eqref{#1}}
\usepackage{algorithm}
\usepackage{amsmath, amsfonts, amssymb} 
\usepackage{graphicx, import} 
\usepackage{bm}
\usepackage{wrapfig}
\usepackage[font=small]{caption}
\usepackage{placeins}
\usepackage{subcaption}
\usepackage{siunitx}
\usepackage{algpseudocode}
\usepackage{cite}
\usepackage{hyperref}
\usepackage{gensymb}
\usepackage[normalem]{ulem}

\newtheorem{definition}{Definition}[section]
\newtheorem{corollary}{Corollary}[section]

\DeclareMathOperator*{\argmin}{argmin}

\newboolean{show_revisions}
\setboolean{show_revisions}{false} 

\newcommand{\revision}[2][]{%
  \ifthenelse{\boolean{show_revisions}}%
    {{\color{red}\sout{#1}}}
    {}
  \ifthenelse{\boolean{show_revisions}}%
    {{\color{blue}#2}}
    {#2}%
}
\title{\LARGE \bf
Safe Low-Altitude Navigation in Steep Terrain\\ with Fixed-Wing Aerial Vehicles
}

\author{Jaeyoung Lim$^{1}$, Florian Achermann$^{1}$, Rik Girod$^{1}$, Nicholas Lawrance$^{2}$, Roland Siegwart$^{1}$
\thanks{Manuscript received: October, 6, 2023; Revised January, 4, 2024; Accepted Jan, 31, 2024. This paper was recommended for publication by Editor Giuseppe Loianno upon evaluation of the Associate Editor and Reviewers’ comments. This work was supported by ETH Research Grant AvalMapper ETH-10 20-1. We would like to thank Yves B\"{u}hler and Elisabeth Hafner at the \ac{SLF} for their expertise and support in avalanche mapping and modeling. We would like to thank Silvan Fuhrer and Thomas Stastny at Auterion AG for supporting the hardware used for the real flight tests.\revision[]{We would also like to thank David Rohr$^{1}$ at ETH Zurich, for being a reliable safety pilot for countless flight tests.}}%
\thanks{$^{1}$ Autonomous Systems Lab, ETH Z\"urich, Z\"urich 8092, Switzerland {\tt \footnotesize \{jalim, acfloria, brik, rsiegwart\}@ethz.ch}}%
\thanks{$^2$ Robotic Perception and Autonomy, CSIRO Data61, QLD 4069, Australia, { \tt\footnotesize nicholas.lawrance@csiro.au}}%
\thanks{Digital Object Identifier (DOI): see top of this page.}
}

\begin{document}

\maketitle

\begin{abstract}
Fixed-wing aerial vehicles provide an efficient way to navigate long distances or cover large areas for environmental monitoring applications. By design, they also require large open spaces due to limited maneuverability. However, strict regulatory and safety altitude limits constrain the available space. Especially in complex, confined, or steep terrain, ensuring the vehicle does not enter \iac{ICS} can be challenging. In this work, we propose a strategy to find safe paths that do not enter \iac{ICS} while navigating within tight altitude constraints. The method uses periodic paths to efficiently classify \acp{ICS}. A sampling-based planner creates collision-free and kinematically feasible paths that begin and end in safe periodic (circular) paths. We show that, in realistic terrain, using circular periodic paths can simplify the goal selection process by making it yaw agnostic and constraining yaw. We demonstrate our approach by dynamically planning safe paths in real-time while navigating steep terrain on a flight test in complex alpine terrain.
\end{abstract}


\section*{SUPPLEMENTARY MATERIAL}
\begin{itemize}
    \item Code: \href{https://github.com/ethz-asl/terrain-navigation}{https://github.com/ethz-asl/terrain-navigation}
    \item Video: \href{https://youtu.be/7C5SsRn_L5Q}{https://youtu.be/7C5SsRn\_L5Q}
\end{itemize}

\section{INTRODUCTION}
\IEEEPARstart{S}{mall} uncrewed aerial systems (\acsp{sUAS}) have become a crucial tool for information-gathering applications such as search and rescue~\cite{oettershagen_robotic_2018}, mapping and inspection~\cite{bircher_three-dimensional_2016}, and environmental monitoring~\cite{jouvet_high_2019, lin_eyewall_2008, shah_multidrone_2020, buhler_photogrammetric_2017}. In particular, fixed-wing and hybrid \ac{VTOL} type vehicles are popular due to their aerodynamic efficiency, resulting in long endurance and extensive area coverage.

Operating fixed-wing vehicles near terrain would enable close-up information-gathering tasks traditionally performed by less efficient multi-rotor vehicles, such as high-resolution near-infrared photogrammetry~\cite{buhler_photogrammetric_2017}. Further, near-surface operations may become necessary as recent regulations require \acp{sUAS} to stay within \SI{120}{\metre} from the closest point of the terrain surface~\cite{eu2019commission}.

Modern fixed-wing \acp{sUAS} have limited online mission planning capabilities, which limit operations to large open areas. Due to the use of aerodynamic forces for maneuvering and power constraints, fixed-wing vehicles are limited by minimum turn radius and \ac{RoC}. The limited maneuverability poses a significant challenge in preventing the vehicle from entering \iacf{ICS}, \revision[ especially when operating close to undulated terrain
. \acp{ICS} are regions of the state space where the vehicle cannot avoid a collision despite all feasible control inputs]{ which are regions of the state space where all control inputs eventually will result in a collision}. \acp{ICS} are more likely to occur in confined environments, e.g. flying below \SI{120}{\metre} in mountainous regions, with terrain steeper than the maximum \ac{RoC}.\revision[U]{An u}nsafe flight can be challenging for a human operator to correct since the vehicle can enter \iac{ICS} long before the event of a collision. Addressing these issues requires an autonomous mission planner that can ensure safety when operating fixed-wing vehicles near steep terrain.

\begin{figure}[t]
\centerline{\includegraphics[width=0.8\linewidth]{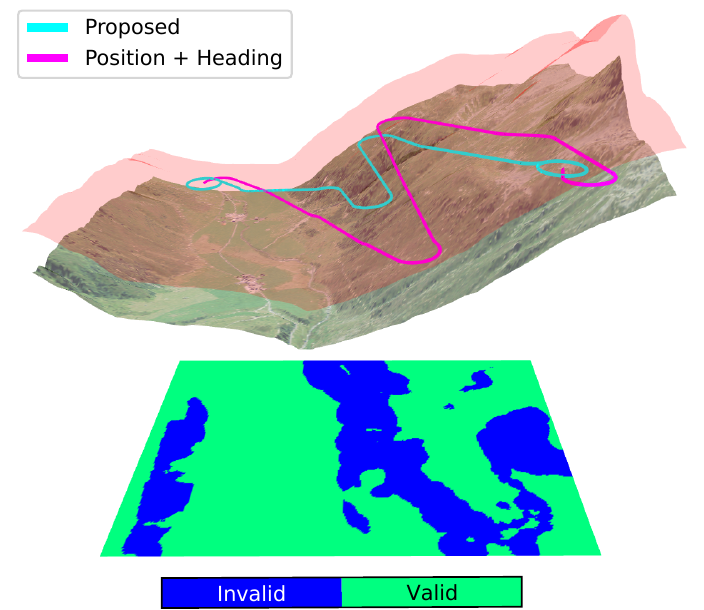}}
\caption{Example of a planned path for a fixed-wing aerial vehicle above complex terrain with an altitude limit of 120 m above the ground (red). The 2D binary projection below the terrain shows the \emph{Valid Loiter Positions} proposed by this paper. Blue regions contain no kinematically feasible safe periodic states. Paths planned using the safe periodic states proposed in this paper (cyan) compared to the conventional goal position and heading (magenta) are shown.}
\label{fig:alpine_planner}
\end{figure}

Previous works have demonstrated practical sampling-based planners for fixed-wing navigation in confined environments~\cite{bry_aggressive_2015, achermann_real-time_2016, oettershagen_towards_2017, lee_optimal_2014, levin_motion_2018, seemann_rrt-based_2014}. 
However, these works only focus on planning collision-free paths between start and goal states without consideration of the \ac{ICS}, significantly reducing their applicability for deployment in altitude-constrained real-world scenarios. This work introduces a \revision[computationally-efficient]{computationally efficient} evaluation of \acp{ICS} using periodic paths for fixed-wing vehicle navigation.\revision[Periodic paths]{Using periodic paths as terminal segments simplifies} the infinite horizon collision checks required to evaluate whether a state is \iac{ICS}. We present circular periodic paths for this particular planning problem, which we pre-compute in the narrow flight corridors above the \ac{DEM}, as seen in \reffig{fig:alpine_planner}. Our approach offloads the operator's workload of selecting a safe goal state by making the goal state yaw agnostic and by further constraining the altitude. Also, the periodicity of terminal paths allows the operator to replan the next mission without time constraints. We demonstrate our approach on a real-world flight navigating a steep valley in the Swiss Alps.

Ultimately, our work enables safe fixed-wing vehicle operation in alpine terrain.\revision[Our planning strategy finds a path that traverses mountainous terrain through valleys and along shallow slopes while respecting terrain distance constraints.]{}The key contributions of this work are:
\begin{itemize}
    \item An efficient method for evaluating \acp{ICS} with sampling-based path planning for fixed-wing navigation.
    \item Evaluation of planner performance on diverse terrains for real-time planning.
    \item A low-altitude flight demonstration in a challenging alpine environment.
    \item An open-source real-time Dubins RRT* planner implementation for fixed-wing \ac{sUAS}.
\end{itemize}

\section{Related Work}
\label{sec:reated_work}
Previous works have explored practical approaches for navigating confined environments with fixed-wing vehicles~\cite{bry_aggressive_2015, levin_motion_2018, achermann_real-time_2016, oettershagen_towards_2017, lee_optimal_2014, seemann_rrt-based_2014}. \cite{bry_aggressive_2015} demonstrated a fixed-wing vehicle operating in confined indoor environments using online trajectory optimization with Dubins polynomials. \cite{levin_motion_2018} use motion primitives to plan acrobatic maneuvers to navigate in cluttered environments. However, trajectory optimization approaches can be\revision[prone to disturbances due to the uncertainty of the dynamics, especially in the presence of wind.]{sensitive to disturbances, especially in the presence of wind.}

Geometric path planning approaches split the problem into path\revision[creation]{planning} and path following. Path\revision[creation]{planning} uses smooth path representations, to represent the dynamic constraints of the vehicle such as Dubins curves~\cite{achermann_real-time_2016}, Bezier curves~\cite{seemann_rrt-based_2014}, and splines~\cite{lee_optimal_2014}.\revision{A guidance controller is then deployed for following the planned path. Therefore, geometric planning approaches are more robust against disturbances, as the reference path is purely geometric and is stabilized around the path with a guidance controller}~\cite{stastny_flying_2019}.

In this work, we use the Dubins airplane model~\cite{chitsaz_time-optimal_2007}, where~\cite{achermann_real-time_2016, oettershagen_towards_2017} show that planning in a geometric Dubins airplane space~\cite{chitsaz_time-optimal_2007} is an effective solution to incorporate the constrained climb rate and curvature for planning collision-free paths in alpine environments. These works formulate the planning problem of finding a collision-free path from start to goal.
Therefore, the operator is still responsible for selecting a safe goal, i.e., not entering \iac{ICS} on arrival at the final state~\cite{fraichard_inevitable_2004}.\revision[This problem is more challenging for fixed-wing vehicles where the vehicle is never stationary.]{However, as fixed-wing vehicles are never stationary, identifying future collisions can be challenging.}

Our work ensures the complete path, including the start and goal, is safe.
\cite{petti_partial_2005}~showed that a collision-free path between two safe states is always safe. Therefore, besides computing a collision-free path, our planner must verify that the start and goals are safe. Computing \acp{ICS} explicitly is challenging since infinite horizon collision checks are required~\cite{bekris_avoiding_2010}. We take inspiration from works that use periodic paths as terminal states to evaluate safety for emergency maneuvers~\cite{arora_emergency_2015} or robust invariant sets~\cite{althoff_online_2015}. We use circular loiter patterns, a typical periodic flight mode for fixed-wing vehicles,\revision[.]{to make an approximate \ac{ICS} verification more efficient.}

\section{PROBLEM FORMULATION}
\label{sec:problem_formulation}
We consider the problem of finding the shortest path $\bm{\eta}^*(s) : \mathbb{R}^+ \rightarrow \mathcal{X}$, where $s$ is the length along the path between any pair of elements from given sets of start and goal states $\mathcal{X}_{start}, \mathcal{X}_{goal}\subset\mathcal{X}$. The complete workspace $\mathcal{X}$ is divided into permitted or free-space states $\mathcal{X}^+ \subseteq \mathcal{X}$, and occupied or prohibited space $\mathcal{X}^- := \mathcal{X} \setminus \mathcal{X}^+$, which are in collision. The permitted workspace $\mathcal{X}^+$ is further partitioned into safe states $\mathcal{X}_{safe} \subseteq \mathcal{X}^+$ and inevitable collision states $\mathcal{X}_{ICS} := \mathcal{X}^+ \setminus \mathcal{X}_{safe}$.

The safe planning problem can be formally written as
\ifthenelse{\boolean{show_revisions}}%
{\textcolor{red}{
\begin{align}
    \bm{\eta}^{*}(s) &= \argmin_{\bm{\eta}} \int_{0}^{S} \Dot{\bm{\eta}}(s) ds \nonumber\\
    \mathrm{s.t.} \quad \bm{\eta}(s) &\in \mathcal{X}_{safe}, &\forall s \in [0, S], \nonumber\\
    \Dot{\bm{\eta}}(s) &= f(\bm{\eta}(s)) &\forall s \in [0, S], \nonumber\\
    \bm{\eta}(0) &\in \mathcal{X}_{start}, \quad \bm{\eta}(S) \in \mathcal{X}_{goal}, \nonumber
\end{align}
}}{}
\revision[]{\begin{align}
    \bm{\eta}^{*}(s) &= \argmin_{\bm{\eta}} \int_{0}^{S} \frac{\partial \bm{\eta}(s)}{\partial s} ds \\
    \mathrm{s.t.} \quad \bm{\eta}(s) &\in \mathcal{X}_{safe}, &\forall s \in [0, S], \nonumber\\
    \frac{\partial \bm{\eta}(s)}{\partial s} &= f(\bm{\eta}(s)) &\forall s \in [0, S], \nonumber\\
    \bm{\eta}(0) &\in \mathcal{X}_{start}, \quad \bm{\eta}(S) \in \mathcal{X}_{goal}, \nonumber
\label{eq:problem_definition}
\end{align}}

where $S$ denotes the path length. The complete path must only contain states in the set of safe states $\mathcal{X}_{safe}$. Also, the path should be kinematically feasible, satisfying the kinematic constraints $f(\cdot)$.

\subsection{Dubins airplane model} \label{sec:problem:dubins_model} 
\begin{figure}[b]
\centerline{\includegraphics[width=0.7\linewidth]{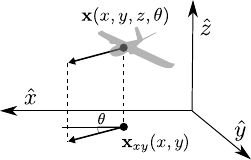}}
\caption{State space of Dubins airplane model.}
\label{fig:dubins_airplane}
\end{figure}
The Dubins airplane~\cite{chitsaz_time-optimal_2007} describes our kinematic model, where the aircraft state consists of a three-dimensional position and heading angle $\bm{x}=\{x,y,z,\theta\}\in \mathbb{R}^3 \times \mathrm{SO}(\revision[1]{2})$, as illustrated in~\reffig{fig:dubins_airplane}. 
We adopt the kinematic definitions from~\cite{chitsaz_time-optimal_2007} to a time-independent form based on distance along the path $\mathbf{s}$. Given a Dubins airplane path $\bm{\eta}(s)$ the path can be reparameterized as
\refequ{eq:dubins_aircraft}, where \revision[$\gamma = \tan^{-1}\left(\dot{z}/\sqrt{\dot{x}^2 + \dot{y}^2}\right)$]{$\gamma = \tan^{-1}\left(\frac{\partial z}{\partial s}/\sqrt{\frac{\partial x}{\partial s}^2 + \frac{\partial y}{\partial s}^2}\right)$} is the flight path angle, and $\kappa$ is the curvature.

\ifthenelse{\boolean{show_revisions}}%
{\textcolor{red}{
\begin{equation}
    \begin{split}
    \Dot{\bm{x}} = f(\bm{x}) = \frac{\partial\bm{x}}{\partial s} = \begin{pmatrix}\Dot{x}\\\Dot{y}\\\Dot{z}\\ \Dot{\theta}\end{pmatrix} =
        \begin{pmatrix}\cos(\gamma)\cos(\theta) \\
         \cos(\gamma)\sin(\theta) \\
         \sin(\gamma) \\
        \kappa \cos(\gamma)  \end{pmatrix}
    \end{split}
    \tag{\ref{eq:dubins_aircraft}}
\end{equation}}}{}
\revision[]{\begin{equation}
    \begin{split}
    \frac{\partial\bm{\eta}(s)}{\partial s}  = f(\bm{\eta}(s))
    = \begin{pmatrix}\cos(\gamma)\cos(\theta)\\
         \cos(\gamma)\sin(\theta) \\
         \sin(\gamma) \\
        \kappa \cos(\gamma)  \end{pmatrix}
    \end{split}
    \label{eq:dubins_aircraft}
\end{equation}
}

The minimum and maximum flight path angle $\gamma \in [\gamma_{min}, \gamma_{max}]$ and path curvature limits $\kappa \in [-\kappa_{max}, \kappa_{max}]$ constrain the maneuverability of the Dubins airplane.

\subsection{Map Representation}
We use a multi-layered surface map~\cite{triebel_multi_2006}, which consists of multiple layers of 2.5D elevation maps to represent the terrain surface and offset collision surfaces. Elevation maps are widely used in robotics~\cite{fankhauser_universal_2016} and geographic information systems communities~\cite{polidori2020digital} to represent large environments.\revision[]{Note that the use of elevation maps might limit the application of representing environments with complex geometry, e.g. indoor or cluttered industrial settings~\cite{triebel_multi_2006}.}

An elevation map $H:\mathcal{M} \mapsto \mathbb{R}$ maps a grid cell position $\bm{m} \in \mathcal{M}$ to the terrain height $h\in \mathbb{R}$, where $\mathcal{M} \subset \mathbb{R}^2$ is the set of all grid cell positions.
\begin{equation}
   h = H(\bm{m}), \quad \bm{m} \in \mathcal{M}
\label{eq:map_representation}
\end{equation}
The planner aims to generate a path in the band between a specified minimum and maximum distance to the closest point on the surface. We define the offset collision surface to represent these constraints.

\begin{definition}[Offset Collision Surface]
The offset collision surface $D_d(\cdot)$ is an elevation map that defines a surface above the terrain where the shortest Euclidean distance to the terrain is $d$ for all points on the surface.
\begin{align}
   D_d(\bm{m}) = \min z & \\
    s.t. \quad \forall \bm{m}_{j} \in \mathcal{M} &\quad \sqrt{(\bm{m} -  \bm{m}_j)^2 + (z - H(\bm{m}_j))^2} \geq d, \nonumber\\
    &\quad z > H(\bm{m}_j) \nonumber
\label{def:distance_surface}
\end{align}
\end{definition}

 We denote the offset collision surface with minimum distance as $D^-$ and maximum distance as $D^+$. A state $\bm{x}=(x, y, z, \theta)$ is considered colliding if the height $z$ is above the maximum offset collision surface $D^+$ or below the minimum offset collision surface $D^-$ at $\bm{x}_{xy}$, where $\bm{x}_{xy}$ is the projection of the state $\bm{x}$ onto the $xy$ plane.
\begin{align}
    \mathcal{X}^- = \{\mathbf{x} \; | \; z >D^+(\mathbf{x}_{xy}) \; \mathrm{or} \; z <D^-(\mathbf{x}_{xy})\}
\end{align}
 
\section{Safe Periodic Sets}
\label{sec:safe_goal_states}
\subsection{Inevitable Collision States}
\revision[The \ac{ICS} concept was ]{\acp{ICS} were }first proposed in~\cite{fraichard_inevitable_2004}, defining a set of states where the vehicle will inevitably result in a collision regardless of all feasible inputs. We adapt the original definition, which uses trajectories, to operate on kinematically feasible paths.
\begin{definition}[Inevitable Collision State\cite{fraichard_inevitable_2004}]
Define $\Gamma(\mathbf{x}, S)$ as the set of all kinematically feasible paths starting from initial state $\mathbf{x} \in \mathcal{X}^+$ with length $S$. $\bm{\eta} \in \Gamma(\mathbf{x}, S)$ is a single path starting from $\mathbf{x}$.
An inevitable collision state is defined as
\begin{align}
    \mathcal{X}_{ICS} = \{\mathbf{x} \; | \; \forall \bm{\eta} \in \Gamma(\mathbf{x}, \infty) \quad \exists s \in [0, \infty) \quad \bm{\eta} (s) \in \mathcal{X}^- \}.\nonumber
\end{align}
\label{def:ics}
\end{definition}
In simple terms, if all possible paths from $\mathbf{x}$ will eventually result in passing through a forbidden state, then $\mathbf{x} \in \mathcal{X}_{ICS}$. Conversely, if there exists a kinematically feasible, collision-free path from $\mathbf{x}$, then $\mathbf{x}$ is not \iac{ICS}.

\subsection{Safe Periodic Path}
An infinite horizon path must be evaluated to ensure a particular state is not in $\mathcal{X}_{ICS}$~\cite{bekris_avoiding_2010}. Since evaluating an infinite horizon path is typically impractical, we use periodic paths to find a subset of all possible safe states $\mathcal{X}_{safe}$.
A path is periodic if any state on the path is repetitively visited with a certain period $\lambda$ as in the following condition:
\begin{align}
    \forall s \in [0, \infty) \quad \bm{\eta}(s) = \bm{\eta}(s+\lambda) \nonumber
\end{align}

\revision[]{The key idea is that if we can find a collision-free path to a periodic state, the path is safe. } If a state $\bm{x}$ is on a periodic path $P(\cdot)$, it is sufficient to check the collision of one period of the path to check whether the state is safe~\cite{bekris_avoiding_2010}.

\begin{corollary}[Safe Periodic Path]
If a periodic path $P(\cdot)$ is not in a collision within a single period, all its states are not in \iac{ICS}.\label{cor:periodic_path}
\begin{align}
    \forall s \in [0, \lambda)\quad P(s) \not \in \mathcal{X}^{-} \iff \forall s \in [0, \infty)\quad P(s) \not \in \mathcal{X}_{ICS} \nonumber
\end{align}
\end{corollary}

While the method is applicable for all periodic paths, we choose a circular path\revision[~to simplify collision checking.]{s, as the isotropic distance to the center of the loiter simplifies collision checking.} Consider a level circular path parameterized by the center position $\bm{c} = (c_x,c_y,c_z) \in \mathbb{R}^3$ and radius $R$. The heading $\theta$ is omitted for brevity.
\begin{align}
    P(\bm{c}, R) = \left\lbrace \bm{x} \; | \; z=c_z , \| \bm{x}_{xy} - \bm{c}_{xy}\| = R\right\rbrace \nonumber
\end{align}

To determine whether the circular path is collision-free, we check whether the states on the path lie between the two offset collision surfaces, $D^+$ and $D^-$. Therefore, a state $\bm{x}$ on a circular path $P(\bm{c}, R)$ is safe if it satisfies the condition~\refequ{eq:is_incollision}.
\begin{equation}
   \forall \bm{x} \in P(\bm{c}, R) \quad D^+_d(\bm{x}) > x_z >D^-_d(\bm{x}).
   \label{eq:is_incollision}
\end{equation}

\begin{figure}[b]
    \centerline{\includegraphics[width=\linewidth]{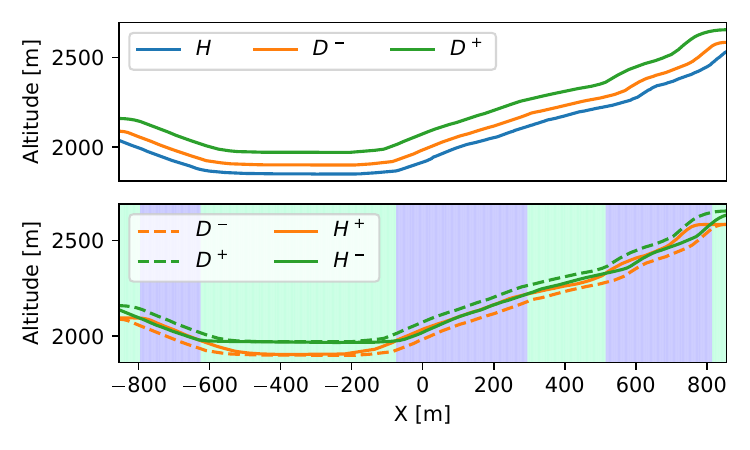}}
\caption{Visualization of the terrain and maximum(\SI{120}{\metre}), minimum distance(\SI{50}{\metre}) offset collision surfaces of the example of~\reffig{fig:alpine_planner} (top). Valid loiter positions according to~\refequ{eq:valid_position} are marked as green and invalid regions are marked as blue (bottom) }
\label{fig:distance_surfce}
\end{figure}

\subsection{Efficient Safety Evaluation}
In~\refequ{eq:is_incollision}, we must iterate through all position states along the circular path for collision-checking. We simplify the computation by replacing the circle with a disk and comparing two horizontal offset surfaces instead. The underlying sufficient condition of a collision-free circular path is to check whether the horizontal distance from the circle’s center to the offset collision surface $D^+$, $D^-$ is larger than the path radius. Thus, we offset $D^+$ and $D^-$ horizontally by radius $R$, resulting in the horizontal offset surfaces $\mathcal{H}^+_R(\bm{m})$ and $\mathcal{H}^-_R(\bm{m})$.
\begin{align}
    \bm{m} \in \mathcal{M} &\quad\mathcal{H}^-_R(\bm{m}) = \max_{\forall \bm{m}' \in \{ \bm{m}' | \|\bm{m}' - \bm{m}\| \leq R\} }D^-(\bm{m}')\nonumber\\
    &\quad\mathcal{H}^+_R(\bm{m}) = \min_{\forall \bm{m}' \in \{ \bm{m}' | \|\bm{m}' - \bm{m}\| \leq R\} }D^+(\bm{m}')
    \label{eq:loiter_position_surface}
\end{align}

If the altitude of the center position $\mathbf{c}=(c_x,c_y,c_z)$ satisfies $c_z > \mathcal{H}^-(\bm{c}_{xy})$ and $c_z < \mathcal{H}^+(\bm{c}_{xy})$, a circular path $P(\mathbf{c}, R)$ is safe. We define the set of \emph{valid loiter positions} over the whole map as follows.
\begin{definition}[Valid Loiter Position]
A collision-free circular path exists at $\bm{m}$ if $\mathcal{H}^{+}(\bm{m}) > \mathcal{H}^{-}(\bm{m})$.
\begin{align}
    \mathcal{M}_{valid} = \{\mathbf{m}\in\mathcal{M}| \mathcal{H}^{+}(\bm{m}) > \mathcal{H}^{-}(\bm{m})\}
    \label{eq:valid_position}
\end{align}
\label{def:valid_loiter_position}
\end{definition}

Following the definition, the permitted loiter positions, used to set the start and goal, are deterministic and can be precomputed for a fixed loiter radius and terrain map. \reffig{fig:distance_surfce} shows that valid loiter positions exist only in flatter terrain.

\section{Path Planning}
\label{sec:planning}

In section~\ref{sec:safe_goal_states}, we showed how the safety of a periodic path can be evaluated using simple geometric operations. This section explains how we find safe routes using sampling-based planners and safe periodic sets.
\revision[]{The information flow of the planning approach is shown in~\reffig{fig:circle_goal}}

\subsection{Start and Goal States}
\label{sec:start_and_goal_states}
A sufficient safety condition is that if a path $\bm{\eta}(s)$ is collision-free and the final state is not in $\mathcal{X}_{ICS}$, the whole path is not in $\mathcal{X}_{ICS}$~\cite{petti_partial_2005}. Therefore, if we can evaluate the safety of $\mathcal{X}_{goal}$ using the strategy proposed in section~\ref{sec:safe_goal_states}, we can solve the problem of finding a collision-free path $\bm{\eta}(s)$ from $\mathcal{X}_{start}$ to $\mathcal{X}_{goal}$. \revision[]{We assume that $\mathcal{X}_{start}$ is safe.}

From~\refequ{eq:loiter_position_surface}, a horizontal goal position is valid if its circle center $\bm{c}_{xy} \in \mathcal{M}_{valid}$.
A valid altitude of the circle center $\bm{c}$ is any value between the two offset surfaces $\mathcal{H}^+_R(\cdot)$, $\mathcal{H}^-_R(\cdot)$. We simplify the altitude selection by choosing the average between the two surfaces, as shown in~\refequ{eq:safe_goal_position_surface}.
\begin{align}
    c_z = \frac{\mathcal{H}^+(\bm{c}_{xy}) + \mathcal{H}^-(\bm{c}_{xy})}{2}
    \label{eq:safe_goal_position_surface}
\end{align}

This results in choosing the circular path with the maximum clearance from the offset collision surfaces. Also, the automatic altitude selection offloads the operator's workload. The valid loiter position check and automatic altitude selection reduce the operation complexity from selecting a four-dimensional goal position and heading $\{x,y,z,\theta\}$ to a two-dimensional circle center selection, guaranteeing safety.

We generate the start and goal sets $\mathcal{X}_{start}$, $\mathcal{X}_{goal}$ by discretizing the circular path $P(\mathbf{c}, R)$ into $N$ discrete sets of states $\hat{P}(\mathbf{c}, R)$, for $i \in \lbrace 1, \cdots,N \rbrace$, as described in~\refequ{eq:circular_goals}.

\begin{figure}[b]
\vspace*{0.2cm}
\centerline{\includegraphics[width=\linewidth]{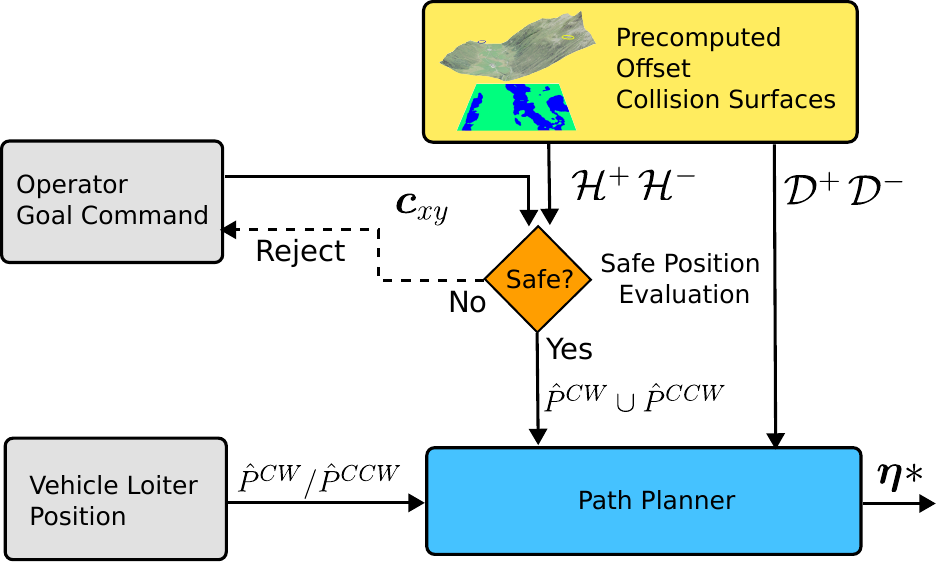}}
\caption{\revision[]{Overview of the proposed method. The operator commands a 2D goal position $c_{xy}$, in which the safety is evaluated using the offset collision surface $\mathcal{H}^+, \mathcal{H}^-$. If the commanded goal position is safe, a planning problem is solved for path $\eta*$ using the start and goal circular path and the collision surface $\mathcal{D}^+, \mathcal{D}^-$}}
\label{fig:circle_goal}
\end{figure}

\ifthenelse{\boolean{show_revisions}}%
{\textcolor{red}{\begin{align}
    \hat{P}(\mathbf{c}, R)
    =\{ \bm{x}_i = \begin{pmatrix}
     R cos(\theta_i) + c_x\\
     R sin(\theta_i) + c_y\\
     c_z \\
     \theta_i 
    \end{pmatrix} | \theta_i = (\frac{i}{N} \pm \frac{1}{2})\pi\}\nonumber
\end{align}}}{}
\revision[]
{\begin{gather}
    \label{eq:circular_goals}
    \hat{P}^{CW}(\mathbf{c}, R)=\{ \bm{x}_i | \theta_i = (\frac{i}{N} + \frac{1}{2})\pi\}\\
    \hat{P}^{CCW}(\mathbf{c}, R)=\{ \bm{x}_i | \theta_i = (\frac{i}{N} - \frac{1}{2})\pi\}\\
    s.t. \quad \bm{x}_i = \begin{pmatrix}
     R cos(\theta_i) + c_x\\
     R sin(\theta_i) + c_y\\
     c_z \\
     \theta_i 
    \end{pmatrix}\nonumber
\end{gather}
}
\revision[While the goal loiter direction is bidirectional, the start loiter direction is fixed since the vehicle is already loitering in a certain direction, as illustrated in \reffig{fig:circle_goal}. Counter-clockwise paths would satisfy $\theta_i = \frac{i}{N} \pi + \frac{\pi}{2}$, clockwise paths satisfy $\theta_i = \frac{i}{N} \pi - \frac{\pi}{2}$. For bidirectional sets, the union of the two goal sets is taken.]{Where $\hat{P}^{CW}(\mathbf{c}, R), \hat{P}^{CCW}(\mathbf{c}, R)$ are clockwise and counter-clockwise circular paths. While the start loiter direction would depend on the vehicle state, both loiter directions are valid for the goal. We define the goal set as the union of both directions.}

\ifthenelse{\boolean{show_revisions}}%
{
\begin{figure}[t]
\centerline{}
\vspace*{0.2cm}
\begin{subfigure}{0.5\linewidth}
\centerline{\includegraphics[width=0.8\linewidth]{figures/unidirectional_goal.pdf}}
    \caption{Unidirectional (clockwise) goal set}\label{fig1:a} \par
\end{subfigure}%
\begin{subfigure}{0.5\linewidth}
\centerline{\includegraphics[width=0.8\linewidth]{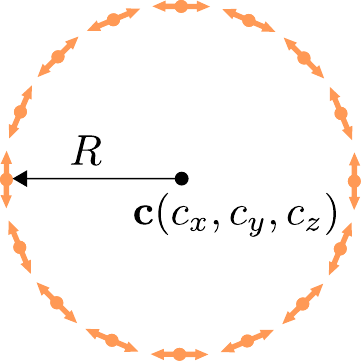}}
    \caption{Bidirectional goal set}\label{fig1:b} \par 
\end{subfigure}%
\caption{\revision[Example of the unidirectional and bidirectional start and goal samples]{}}
\end{figure}
}{}

\begin{figure*}[t]
\begin{subfigure}{0.33\textwidth}
    \includegraphics[width=\linewidth]{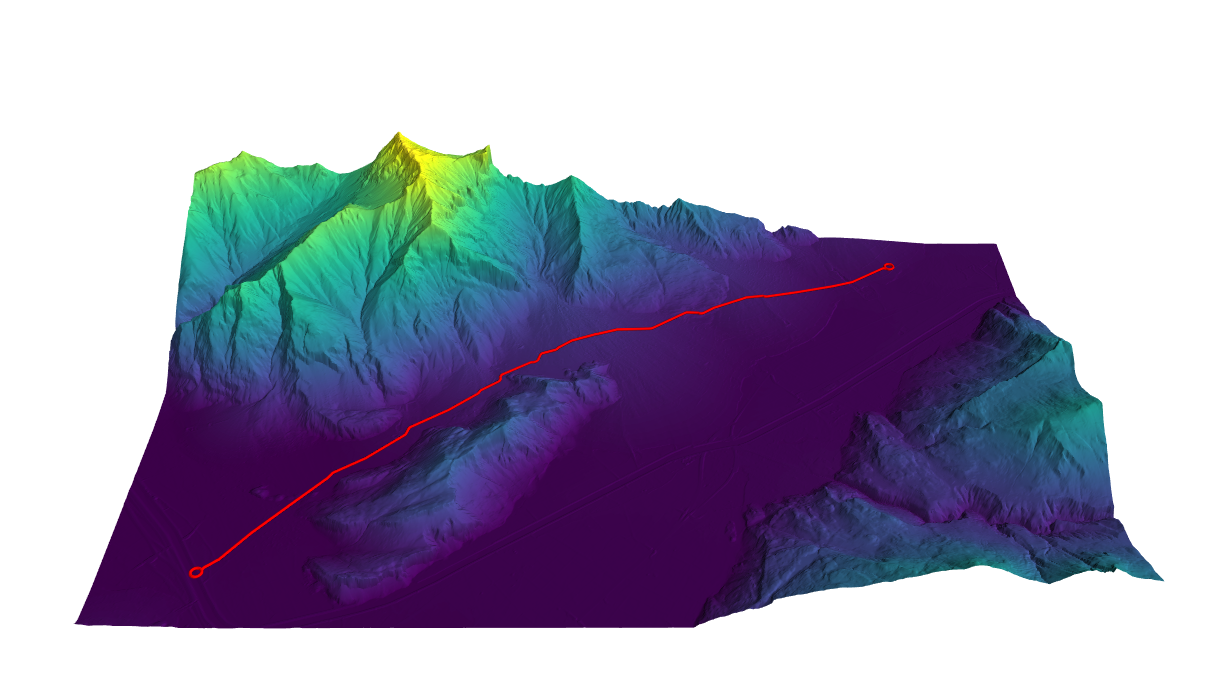}
    \includegraphics[width=\linewidth]{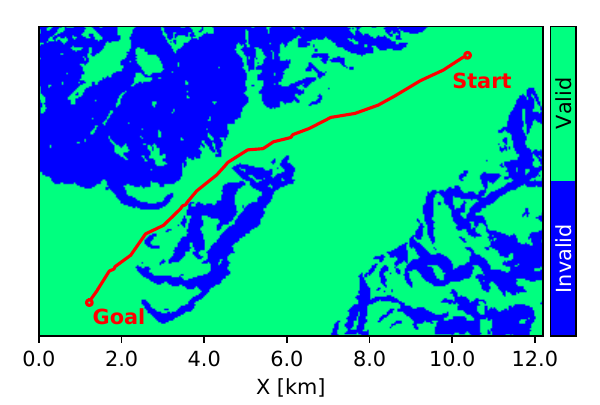}
    \caption{\emph{Sargans}}
    \label{fig:evaluation_terrain:a}
\end{subfigure}
\begin{subfigure}{0.33\textwidth}
    \includegraphics[width=\linewidth]{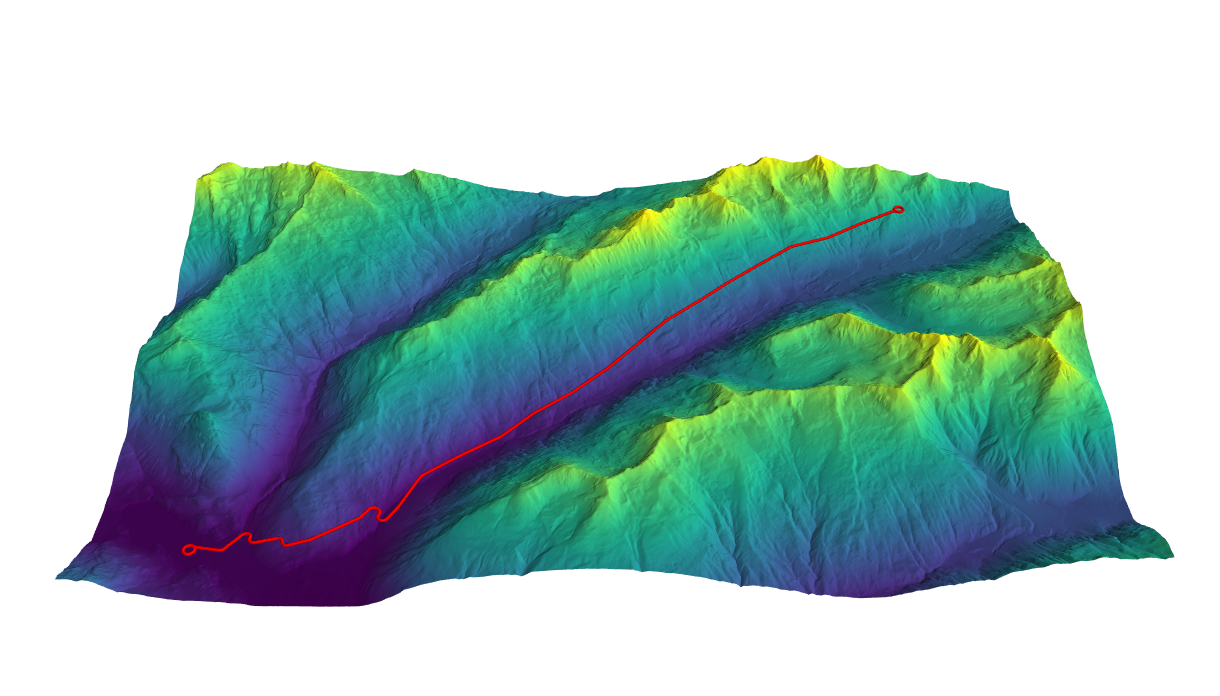}
    \includegraphics[width=\linewidth]{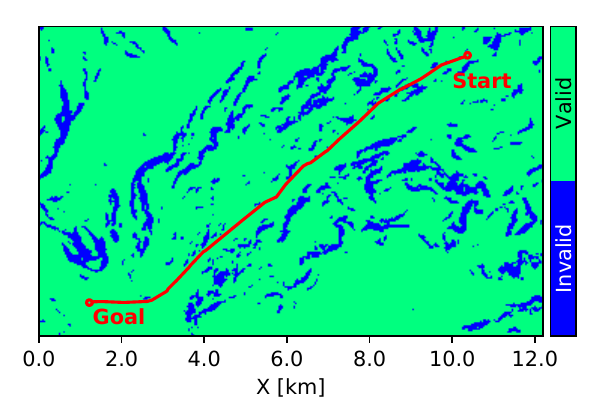}
    \caption{\emph{Dischma Valley}}\label{fig:evaluation_terrain:b} \par 
\end{subfigure}%
\begin{subfigure}{0.33\textwidth}
    \includegraphics[width=\linewidth]{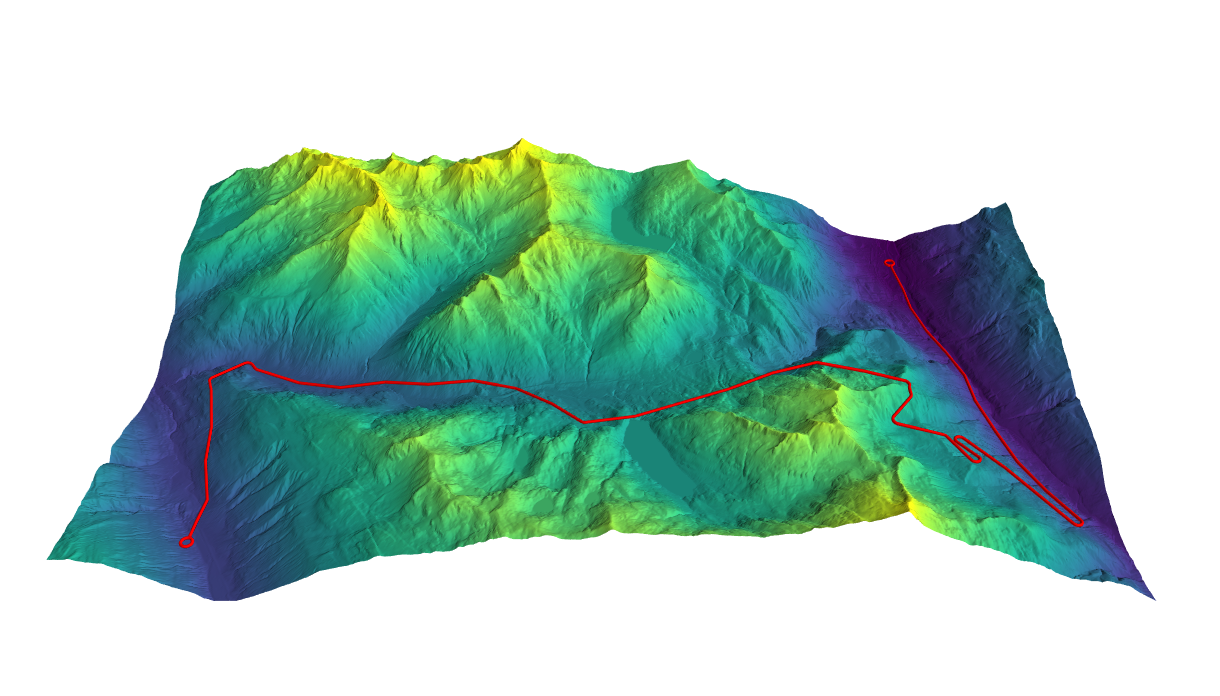}
    \includegraphics[width=\linewidth]{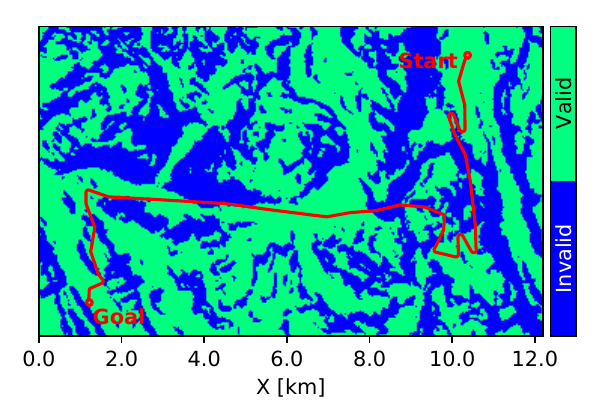}
    \caption{\emph{Gotthard Pass}}\label{fig:evaluation_terrain:c} \par
\end{subfigure}%
\caption{(above) 3D visualization of the elevation data and an example of a planned path and (below) two-dimensional projection of the valid loiter position set $\mathcal{M}_{valid}$ (green) in three environments: (a) \emph{Sargans}, (b) \emph{Dischma Valley}, and (c) \emph{Gotthard Pass}.}
\label{fig:evaluation_terrain}
\end{figure*}

\subsection{Planning}
We plan using RRT*, a probabilistically complete, asymptotically optimal sampling-based planner~\cite{karaman_incremental_2010}.
The planner samples in the Dubins airplane space, \revision[where we define a space with a quasi-distance metric $d_D(x_i, x_j)$ such that $d_D$ represents the length of the shortest feasible path]{which is defined by a quasi-distance metric $d_D(x_i, x_j)$ such that $d_D$ is the length of the Dubins airplane path between two arbitrary states $x_i, x_j \in \mathcal{X}$}~\cite{chitsaz_time-optimal_2007}. \revision[]{In this work, we use the minimum turn radius to be identical with the terminal circular path radius explained in~\refsec{sec:start_and_goal_states}, since using the minimum turn radius maximizes valid loiter positions. However, the radius of the terminal circular path can be set independently of the minimum turn radius of the Dubins airplane space.} Note that the distance metric is a quasi-metric, given that the distance may not be symmetric between the two states $x_i, x_j$.

To calculate the shortest path, we use~\cite{mclain_implementing_2015}, which divides the Dubins distance in the horizontal plane and altitude differences into three cases to determine whether an ascending/descending helix is required to reach the goal while satisfying the flight path angle constraints. As in~\cite{schneider_path_2016}, we take the sub-optimal Dubins distance to remove the need for a line search in the \emph{medium goal altitude} case. To make the Dubins curve calculation more efficient, we use the set classification method proposed in~\cite{shkel_classication_2001} to prevent exhaustive computation of different Dubins path types. \revision[Since a Dubins curve~\cite{dubins_curves_1957} consists of arcs and straight lines, this results in the path having discrete curvatures such as $\kappa \in \{-\kappa_{max}, 0, \kappa_{max}\}$.]{}

\section{Planner Evaluations}
We run two quantitative evaluations for the proposed approach. First, we illustrate how terrain ruggedness reduces the safe terminal state space and how our approach relaxes this more complex planning problem on an operational level. Second, we show that our planner is capable of onboard real-time path planning, even in mountains with narrow passages.

\subsection{Setup}

The approach is evaluated in three different terrain environments: \emph{Sargans}, \emph{Dischma Valley}, and \emph{Gotthard Pass}. \reffig{fig:evaluation_terrain} shows the selected \acp{DEM} from SwissAlti3D~\cite{swisstopo2023swissalti3d} with the extent of $12.2\si{km} \times 7.48\si{km}$ with \SI{10}{\metre} lateral resolution.\revision[]{The \ac{DEM} uses CH1903/LV03 coordinates, with Bessel 1841 as the vertical datum. }All terrains are identical in extent but have increasingly rugged topography. The \emph{Gotthard Pass} environment is the most difficult to traverse, containing only a narrow pass.
The minimum and maximum distance to the terrain was \SI{50}{\metre} and \SI{120}{\metre}, corresponding to European regulations~\cite{eu2019commission}. The vehicle's minimum turn radius is $R=\qty{66.67}{\m}$. \revision[The start and goal circular paths have the same radius as the minimum turn radius.]{The maximum flight path angle is $\gamma=8.6\degree$.} The planners were implemented in OMPL~\cite{sucan_ompl_2012} and \iac{ROS}. Map representations use the \verb|grid_map| toolbox for efficiently representing 2.5D surfaces~\cite{fankhauser_universal_2016}. We executed all benchmarks on a \SI{2.9}{\giga\hertz} Intel Core i7-10700 CPU with \SI{32}{\giga\byte} memory.

\subsection{Safe Periodic Goal Selection}
A major difficulty when planning fixed-wing missions in mountainous regions is the selection of the goal state heading. A poorly selected heading may lead to crashing into the mountain upon arrival or an overly complicated approach path as illustrated in \reffig{fig:alpine_planner}. Our safe periodic goal selection relaxes this problem by making the terminal state yaw agnostic with periodic paths.

To demonstrate, we compare the reachable subset of the map when selecting a terminal circle with a \emph{Valid Loiter Position}. \reffig{fig:evaluation_terrain} shows the valid \emph{Valid Loiter Position} locations marked green for the three environments. 
Loitering is generally possible in shallow regions, while steep regions prevent loitering due to the narrow space available. This effect becomes more severe as the environment becomes more rugged, effectively making the planning problem harder. The portion of \emph{Valid Loiter Position} location of each environment is \SI{77}{\percent} for \emph{Sargans}, \SI{79}{\percent} for \emph{Dischma Valley}, \SI{73}{\percent} for the \emph{Gotthard Pass}, with the \emph{Gotthard Pass} environment being very jagged. 
Furthermore, the red paths show that the \emph{Valid Loiter Position} selection ensures a smooth goal approach. \revision[No final turn is necessary to reach the goal, as the planner can always select a suitable heading from the circle that is approximately tangential to the shortest route.]{Note that for the \emph{Gotthard Pass} environment, the path needs to pass through invalid loiter positions, meaning that the vehicle would not be able to initiate a loiter safely.}

In summary, the proposed safe periodic goal selection makes the goal selection process simpler and lowers the workload by eliminating the need to specify the heading as a terminal state.

\begin{figure}[t]
    \centerline{\includegraphics[width=\linewidth]{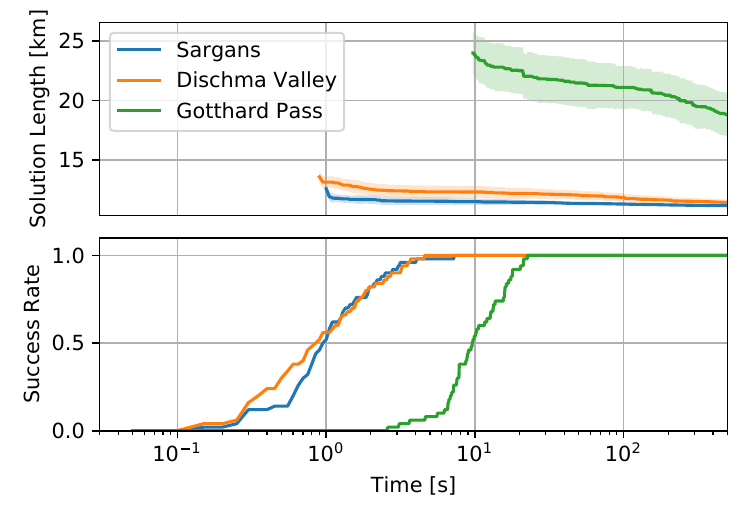}}
    \caption{Planner success rates and solution path length using the proposed planning approach.}
\label{fig:benchmark_planner}
\end{figure}

\subsection{Onboard Planner Convergence}
The proposed safe periodic goal selection provides an opportunity for simpler on-the-fly replanning. We show that our approach can quickly find a valid path onboard the vehicle. For this experiment, we fix the start and goal circle at $\bm{c}_{start} = (2992, -4720, \cdot)[m]$, $\bm{c}_{goal} = (-2992, 4880, \cdot)[m]$ relative to the center of the map for all maps, with both radii of $R=\qty{66.67}{\m}$. We repeat the planning \SI{50}{\times} with a compute time budget of \SI{500}{\second} for each planning loop. The altitude of the start and goal positions are omitted since it is defined by the terrain using~\refequ{eq:safe_goal_position_surface}.

\reffig{fig:benchmark_planner} shows that the planner always finds a solution. The median time to find the initial solution is \SI{1.00}{\second} for \emph{Sargans}, \SI{0.90}{\second} for \emph{Dischma Valley}, and \SI{9.72}{\second} for the more difficult \emph{Gotthard Pass}. A vehicle flying at \SI{15}{\metre\per\second} would take \SI{28}{\second} to fly a full loiter with \SI{66.7}{\metre} radius. Therefore, the vehicle can find a safe path within a fraction of the time required to execute one loiter maneuver.

Despite the start and goal having the same Euclidean distance in all three test cases, the valid path length is significantly longer for \emph{Gotthard Pass}. Also, the solution length does not converge within \SI{500}{\second}. Both of these results are due to the steep mountain faces that make it difficult to connect Dubins airplane path segments, especially in the presence of climb rate, curvature, and altitude constraints. However, the proposed planner always finds a valid initial solution within \SI{30}{\second} in all test cases. \revision{Note that while not converged, the path is always safe and therefore still safe to follow for the vehicle. }Here, the relaxed goal state positively influences the success rate, simplifying goal state connection. \revision[Note that the solution shown in \reffig{fig:evaluation_terrain:c} resembles the road contour from Realp (UR) to Airolo (TI), highlighting the planner's natural behavior to adapt to the mountain slope. ]{However, since the planning approach is identical except for the larger goal set, the difference in convergence was minimal.}

\section{Real World Flight Tests}

We conducted an actual flight experiment on \iac{sUAS}. The primary goal of this experiment was to demonstrate that our planner enables safe and legal flights in steep environments. Secondary, we validate our onboard replanning capabilities. 
\subsection{Setup}
\label{sec:realworld_tests}
\begin{figure}[t]
\vspace*{0.2cm}
\begin{subfigure}{\linewidth}
\centerline{\includegraphics[width=\linewidth]{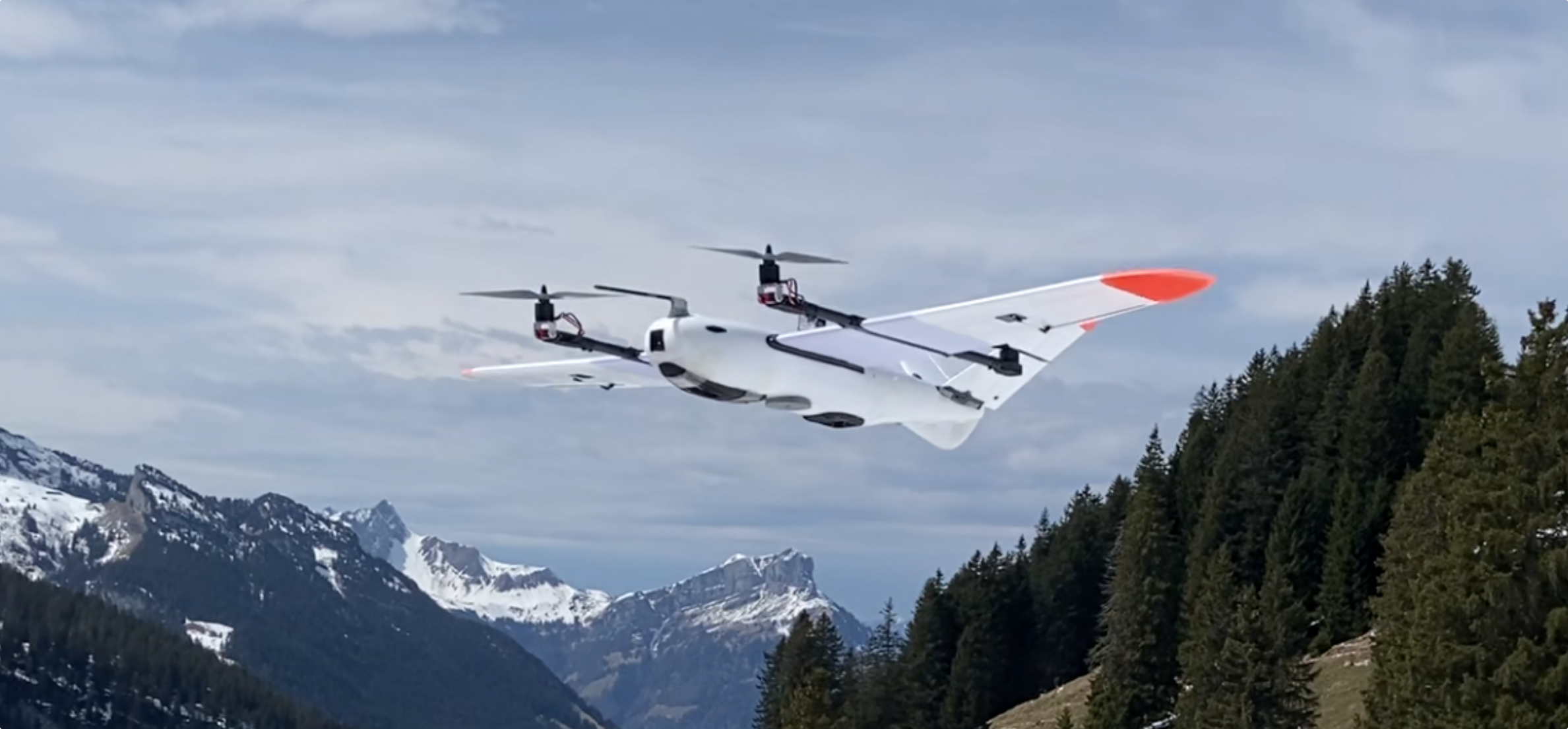}}
\caption{Tiltrotor VTOL platform in flight during take off in Riemenstalden, Switzerland.}
\label{fig:test_vehicle}
\end{subfigure}
\begin{subfigure}{\linewidth}
\centerline{\includegraphics[width=\linewidth]{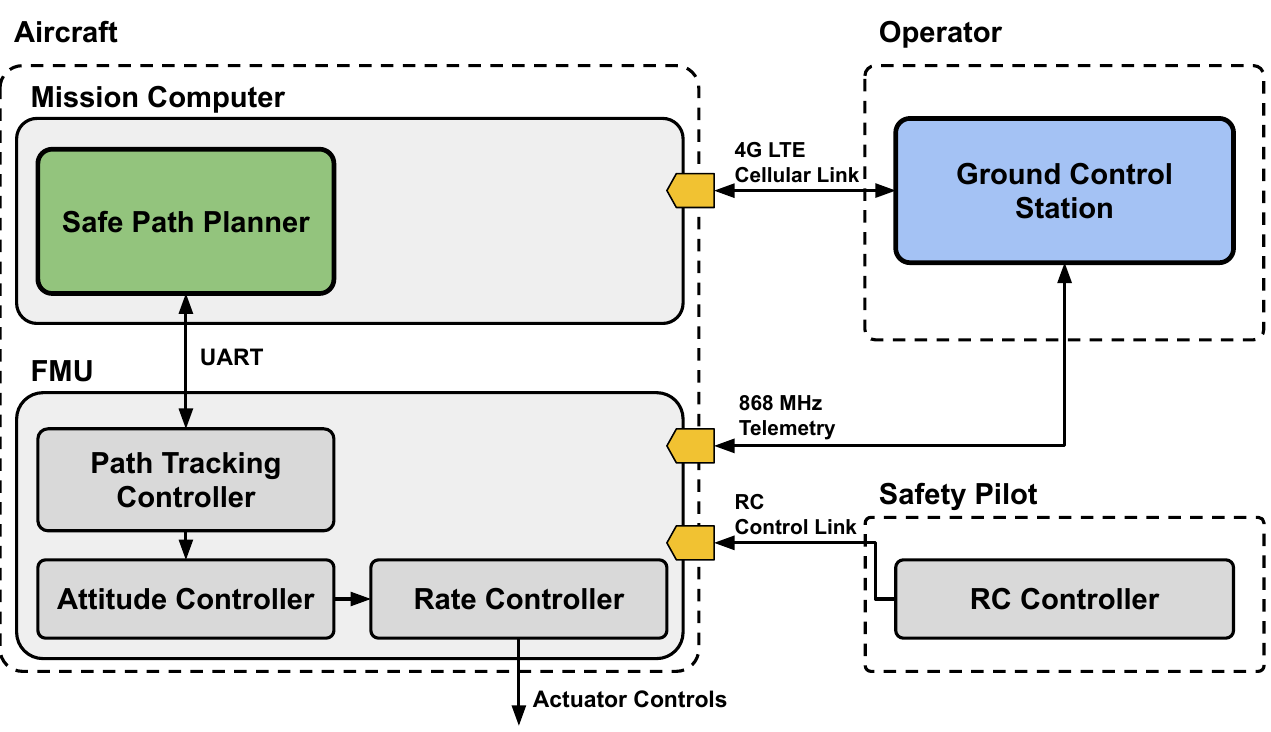}}
\caption{The system consists of a Mission computer, \ac{FMU} which are controlled by an operator and a safety pilot.}
\label{fig:communication_flow}
\end{subfigure}
\caption{Overview of the system that was used for flight testing.}
\label{fig:system_overview}
\end{figure}

\begin{figure*}[t]
\begin{subfigure}[t]{0.37\textwidth}
    \includegraphics[width=\linewidth]{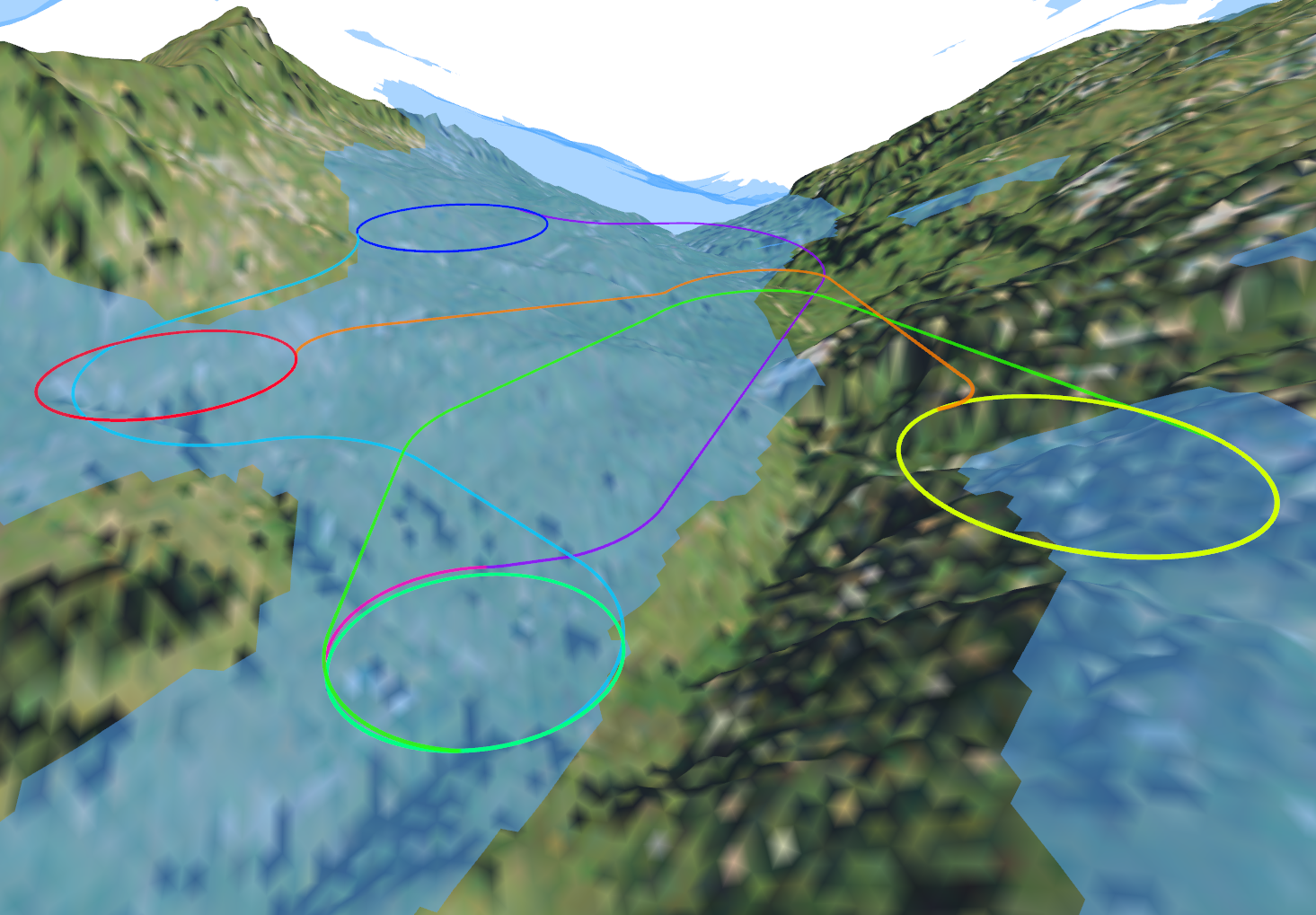}
    \caption{3D visualization of reference path over terrain. }\label{fig:real_flight_test:a} \par
    \label{fig:real_flight_test:a} \par 
\end{subfigure}%
\begin{subfigure}[t]{0.63\textwidth}
    \includegraphics[width=\linewidth]{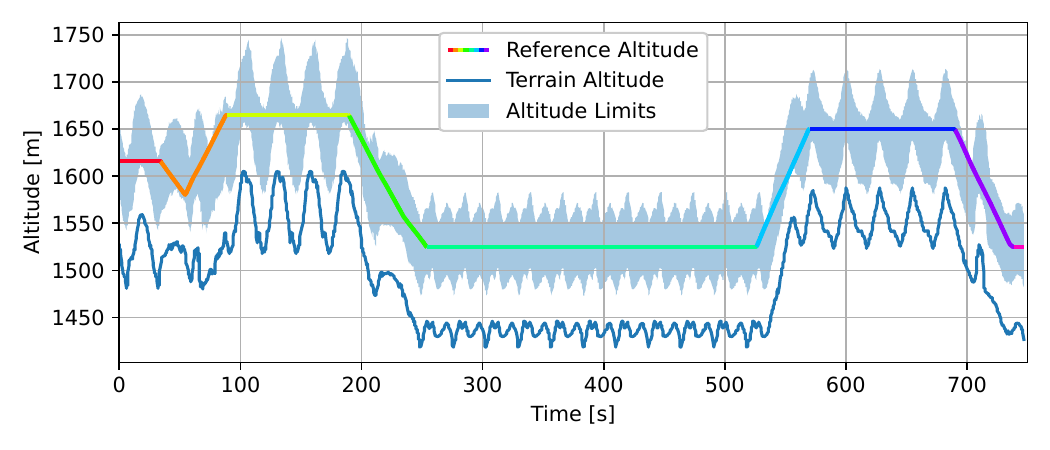}
    \caption{Time record of vehicle altitude, terrain altitude, and offset collision surfaces $D^+$ and $D^-$.}
    \label{fig:real_flight_test:b} \par 
\end{subfigure}%
\caption{Visualization of the vehicle path during flight tests in the valley of Riemenstalden, Switzerland. Mission legs are color-coded.  \emph{Valid loiter position} surfaces are shaded blue. The altitude limits are defined by the offset collision surface $D^+$ and $D^-$. \revision[]{Due to small vertical tracking errors the vehicle altitude overlaps with the reference, thus was omitted from this figure and instead refer to~\reffig{fig:distance_to_terrain}}}
\label{fig:real_experiments}
\end{figure*}
The test was conducted in the valley of Riemenstalden, Switzerland. A video of the demonstration can be found in the supplementary materials. \reffig{fig:real_flight_test:a} shows a terrain excerpt provided by SwissAlti3D~\cite{swisstopo2023swissalti3d}. The terrain is characterized by a relatively steep and narrow valley with a larger flat area at the bottom and a few plateau-like areas on the slopes. In the experiment, the operator dynamically provided four arbitrary goal positions from the safe set shown to the operator through a graphical user interface. The \ac{sUAS} autonomously replanned during loitering and executed the new missions.

The experiment platform was a tiltrotor \ac{VTOL} aircraft with a mass of \SI{5.7}{\kg}, and a wingspan of \SI{2300}{\mm} based on the Makefly Easy Freeman, shown in \reffig{fig:test_vehicle}. The platform hovers during takeoff and landing with the tractor motors in front of the wing tilted upwards. During the remaining flight, the vehicle flies as a fixed-wing vehicle with the front rotors tilted forward. The minimum turning radius of the vehicle is assumed to be \SI{80}{\metre}. The maximum flight path angle is $8.6\degree$. The planner's lower altitude bound was \SI{50}{\metre} to remain safely above the canopy of the trees. According to the legal requirements, the upper altitude bound was \SI{120}{\metre}~\cite{eu2019commission}.

The platform carries an Intel NUC, equipped with a \SI{3.5}{\giga\hertz} Intel Core i7-7567U CPU, which runs the proposed planning method on demand. \reffig{fig:communication_flow} shows the communication flow. 
The operator communicates to the mission computer through a cellular connection. Upon user approval, the mission computer sends setpoints to \iac{FMU} at \SI{10}{\hertz}. The \ac{FMU} runs stock PX4 autopilot with regular \acs{GNSS} navigation. 
The ground station communicates with the autopilot through a cellular connection. For example, telemetry data from the plane is visualized on the \ac{DEM} for planning a new mission. An additional \SI{868}{\mega\hertz} telemetry connection is used for redundancy, and an RC link connects a safety pilot to the vehicle to abort the mission in case of an emergency.

\subsection{Smoothed Dubins Path Tracking}
\label{sec:path_tracking}
The mission computer continuously sends path-tracking reference commands $\mathbf{r} = [\mathbf{p}, \mathbf{v}, \kappa]$ \revision[with the closest point]{from the closest point on the Dubins airplane path }$\mathbf{p}$, tangent $\mathbf{t}$, and curvature $\kappa$ to the \ac{FMU}. The reference is passed to a nonlinear path following guidance controller based on~\cite{stastny_flying_2019}.

Given that Dubins curves consist of arcs and line segments, the curvature is a discrete set, which results in discontinuous jumps in reference curvature. The jumps degrade the tracking performance of the path-tracking controller. Therefore, a linear curvature blending strategy smoothes the curvature reference for better tracking performance. Given the current segment $i$, with curvature $\kappa_i$, and the next segment $i+1$, with curvature $\kappa_{i+1}$, we blend the two segments with the portion $\psi \in [0, 1]$, such that
\begin{align}
    \kappa = \psi \kappa_i + (1-\psi)\kappa_{i+1},
    \label{eq:path_tracking_smoothing}
\end{align}
where $\psi = \max(1.0, l/\Bar{l})$, $l$ is the remaining distance to the end of the segment, and $\Bar{l}$ is the threshold where the segment is blended when nearing the end of the segment. For the experiments, $\Bar{l}$ was \SI{10}{\metre}\revision[]{, which was tuned with respect to the vehicle's roll time constant}.

\subsection{Safe and Legal Flight Test}
\reffig{fig:real_flight_test:a} shows a total of four individual mission legs. The operator commanded all circular goal centers to lie within the safe goal surface. The safe regions are generally located in the flatter parts of the valley, and the vehicle can traverse through steep slopes only for climbing and descending. Exemplary is the transition from the yellow to the green circle. The platform cannot go straight to the goal because of its limited descending rate and tight altitude bounds. Instead, it has to perform a large detour along the slope to reach the valley.
\begin{figure}[t]
\centerline{\includegraphics[width=\linewidth]{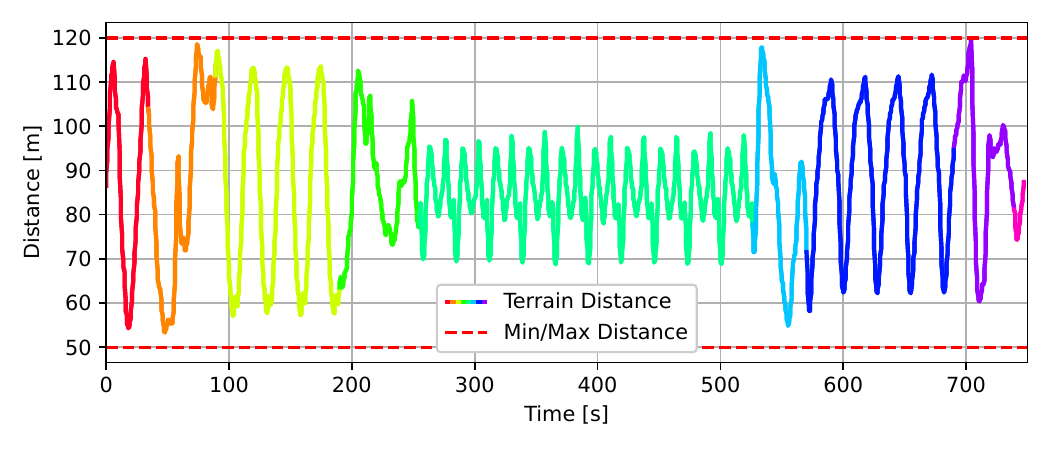}}
\centerline{\includegraphics[width=\linewidth]{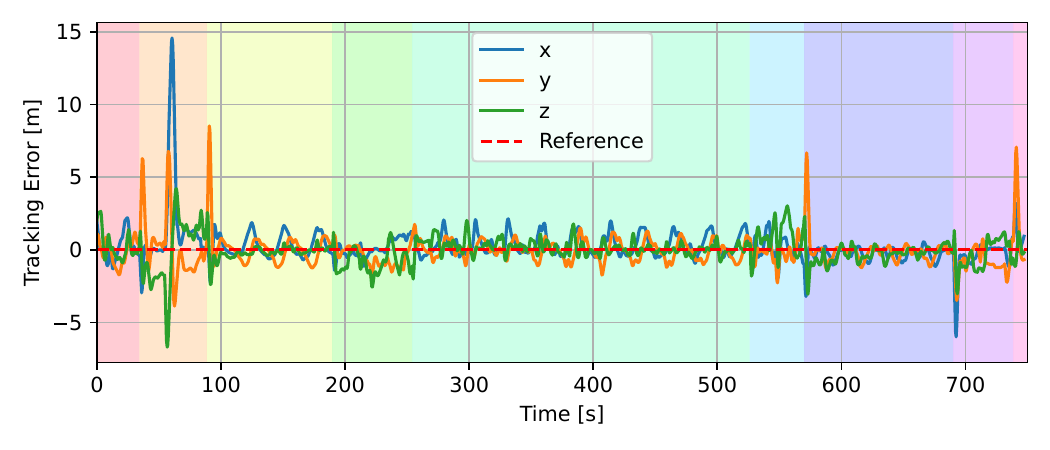}}
\caption{Visualization of the distance of the vehicle to terrain with segment colors (top) and the path tracking error with segment colors overlaid (bottom) during the flight test.}
\label{fig:distance_to_terrain}
\end{figure}

\reffig{fig:real_flight_test:b} gives more insight into the planned path.
The plot reveals that the planned reference path always remains between the minimum and maximum offset collision surface constraints. The path passes as close as \SI{0.56}{\metre} to the maximum distance constraint and \SI{0.915}{\metre} to the minimum distance constraint. Notice the significant periodic variation in altitude limits during the yellow loiter circle between \SIrange{100}{200}{\second}. This is due to the level, circular flight over the steep slope, stressing the necessity of terrain-aware mission planning to remain within altitude limits in non-flat environments.

Finally, we investigate the path tracking error to determine how well the flight controller handles the Dubins path approximation.
\reffig{fig:distance_to_terrain} shows good altitude tracking, where the tracking errors were \SI[separate-uncertainty = true]{0.65(0.73)}{\metre} with a maximum of \SI{6.69}{\metre}. On the other hand, the lateral tracking errors were \SI[separate-uncertainty = true]{1.06(1.37)}{\metre}, with a maximum of \SI{14.6}{\metre}. Both the maximum altitude and lateral tracking errors occur in the middle of the orange segment (\qty{\sim 60}{\s}), where a full right-handed turn and climb occur after a long descent. We attribute these errors to the curvature discontinuity in the Dubins path discussed in \refsec{sec:path_tracking}. Despite the tracking errors, the platform remains within the terrain distance constraints. The larger deviation occurred when the clearance to both distance constraints was large. 

Our experiment shows that for our system and this environment, the planner parametrization was well chosen. In the case of a less performant tracking controller or steeper terrain, the altitude limits could be tightened, the climb rate reduced, or the minimum turning radius enlarged to increase safety at the expense of a smaller workspace. 

\section{Conclusions}
This work investigates autonomous fixed-wing flight in altitude-constrained mountainous regions. Based on the concept of \iacf{ICS}, we propose a safe planning approach that uses circular periodic paths to verify the safety of a goal state. We show that circular periodic paths enable us to simplify the goal selection process, compared to conventional start and goal states specified by a singular position and heading. We incorporate this into a sampling-based planning framework and demonstrate the planner in real-world experiments. The experiments highlight the necessity for terrain-aware mission planning in the mountains, where climb rate, curvature, and altitude constraints drastically limit the search space.

To the authors' knowledge, this is the first demonstration of a fixed-wing \acp{sUAS} autonomously navigating in alpine terrain, complying with the European regulations~\cite{eu2019commission}. \revision{The main limitation of this work is that the path planning is done in the Dubins Airplane space, which is only able to represent a small subset of possible paths. Also, environmental effects such as wind are ignored from the planning problem. }
Future work may include safe, any-time replanning due to prioritized air traffic, handling wind and tracking uncertainty, or deviation from the nominal path. 

\bibliographystyle{IEEEtran}
\bibliography{references_local}

\end{document}